\documentclass[runningheads, vietnamese]{llncs}
\usepackage{graphicx}
\usepackage[utf8]{inputenc}
\usepackage[T5]{fontenc}
\usepackage{varwidth}
\usepackage{adjustbox}
\usepackage{float}

\begin{document}
\title{UIT-ViIC: A Dataset for the First Evaluation on Vietnamese Image Captioning}
\titlerunning{A Dataset for the First Evaluation on Vietnamese Image Captioning}
%
\author{Quan Hoang Lam \inst{1,2} \and
Quang Duy Le\inst{1,2} \and
Kiet Van Nguyen \inst{1,2}  \and
Ngan Luu-Thuy Nguyen\inst{1,2}}
\authorrunning{Lam. et al.}
%
\institute{University of Information Technology, Ho Chi Minh City, Vietnam \and
Vietnam National University, Ho Chi Minh City, Vietnam\\
\email{\{15520673, 15520687\}@gm.uit.edu.vn,\{kietnv, ngannlt\}@uit.edu.vn }}
\maketitle              
\begin{abstract}
Image Captioning, the task of automatic generation of image captions, has attracted attentions from researchers in many fields of computer science, being computer vision, natural language processing and machine learning in recent years. This paper contributes to research on Image Captioning task in terms of extending dataset to a different language - Vietnamese. So far, there is no existed Image Captioning dataset for Vietnamese language, so this is the foremost fundamental step for developing Vietnamese Image Captioning. In this scope, we first build a dataset which contains manually written captions for images from Microsoft COCO dataset relating to sports played with balls, we called this dataset UIT-ViIC. UIT-ViIC consists of 19,250 Vietnamese captions for 3,850 images. Following that, we evaluate our dataset on deep neural network models and do comparisons with English dataset and two Vietnamese datasets built by different methods. UIT-ViIC is published on our lab website \footnote{\url{https://sites.google.com/uit.edu.vn/uit-nlp/}} for research purposes.

\end{abstract}
\section{Introduction}

Generating descriptions for multimedia contents such as images and videos, so called Image Captioning, is helpful for e-commerce companies or news agencies. For instance, in e-commerce field, people will no longer need to put much effort into understanding and describing products' images on their websites because image contents can be recognized and descriptions are automatically generated. Inspired by Horus \cite{horus2017}
, Image Captioning system can also be integrated into a wearable device, which is able to capture surrounding images and generate descriptions as sound in real time to guide people with visually impaired. 

Image Captioning has attracted attentions from researchers in recent years \cite{karpathy2015deep}, \cite{vinyals2015show}, \cite{yoshikawa2017stair}, and there has been promising attempts dealing with language barrier in this task by extending existed dataset captions into different languages \cite{yoshikawa2017stair}, \cite{li2019coco}. 



\begin{figure}
    \centering
    \includegraphics[width=\linewidth]{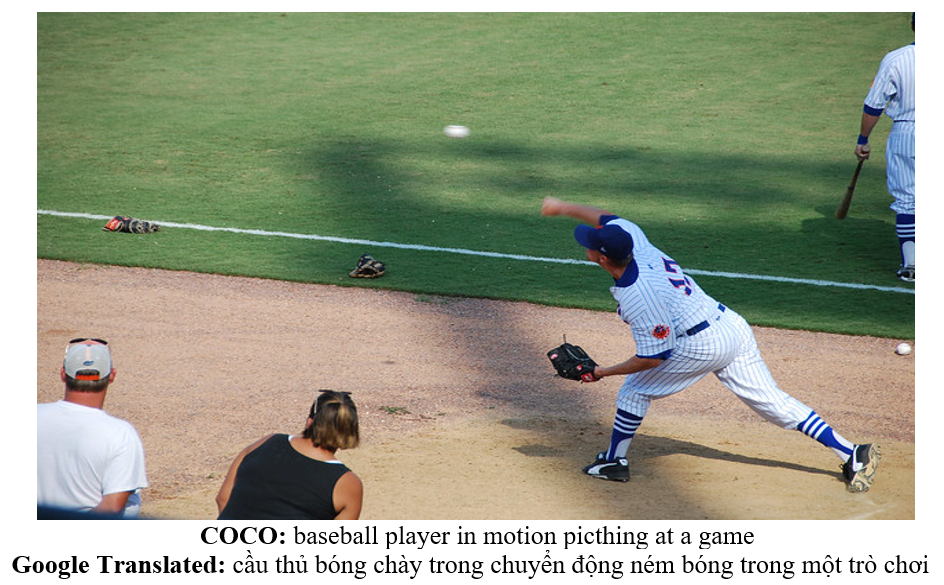}
    \caption{\textbf{Example of MS-COCO English caption compare to Google Translated caption}}
    \label{fig:cultural_comparison}
 \end{figure}

In this study, generating image captions in Vietnamese language is put into consideration. One straightforward approach for this task is to translate English captions into Vietnamese by human or by using machine translation tool, Google translation. With the method of translating directly from English to Vietnamese, we found that the descriptions are sometimes confusing and unnatural to native people. Moreover, image understandings are cultural dependent, as in Western, people usually have different ways to grasp images and different vocabulary choices for describing contexts. For instance, in Fig. \ref{fig:cultural_comparison}, one MS-COCO English caption introduce about "a baseball player in motion of pitching", which makes sense and capture accurately the main activity in the image. Though it sounds sensible in English, the sentence becomes less meaningful when we try to translate it into Vietnamese. One attempt of translating the sentence is performed by Google Translation, and the result is not as expected.
\vspace{-1em}
 \vspace{-1em}
 
Therefore, we come up with the approach of constructing a Vietnamese Image Captioning dataset with descriptions written manually by human. Composed by Vietnamese people, the sentences would be more natural and friendlier to Vietnamese users. The main resources we used from MS-COCO for our dataset are images. Besides, we consider having our dataset focus on sportball category due to several reasons:

\begin{itemize}
    \item By concentrating on a specific domain we are more likely to improve performance of the Image Captioning models. We expect our dataset can be used to confirm or reject this hypothesis.
    \item Sportball Image Captioning can be used in certain sport applications, such as supportting journalists describing great amount of images for their articles. 
\end{itemize}

Our primary contributions of this paper are as follows:

\begin{itemize}
  \item Firstly, we introduce UIT-ViIC, the first Vietnamese dataset extending MS-COCO with manually written captions for Image Captioning.  UIT-ViIC is published for research purposes.
  \item Secondly, we introduce our annotation tool for dataset construction, which is also published to help annotators conveniently create captions.
  \item Finally, we conduct experiments to evaluate state-of-the-art models (evaluated on English dataset) on UIT-ViIC dataset, then we analyze the performance results to have insights into our corpus.
\end{itemize}

The structure of the paper is organized as follows. Related documents and studies are presented in Section \ref{related}. UIT-ViIC dataset creation is described in Section \ref{dataset}. Section \ref{models} describes the methods we implement. The experimental results and analysis are presented in Section \ref{experiment}. Conclusion and future work are deduced in Section \ref{conclusion}.

\section{Related Works}
\label{related}
\vspace{-2em}
\begin{table*}
\caption{Public image datasets with manually annotated, non-English descriptions}
\begin{adjustbox}{width=1\textwidth}
{%
  \begin{tabular}{|l|c|c|c|r|r|c| }
    \hline
    \textbf{Dataset}& \textbf{Release} &\textbf{Data source}&\textbf{Languages} & \textbf{Images} &\textbf{Setences} &\textbf{Application} \\
\hline
IAPR TC-12 &2006 &Internet &English/German&20,000&100,000&Image retrieval\\
\hline
Pascal sentences &2015 &Pascal sentences &Japanese/English &1,000 &5,000 &Cross-lingual document retrieval\\
\hline
YJ Captions &2016 &MS-COCO &Japanese/English &26,500 &131,470 &Image Captioning \\
\hline
MIC test data &2016 &MS-COCO &French/German/English &1,000 &5,000 &Image retrieval \\
\hline
Bilingual caption &2016 &MS-COCO &German/English &1,000 &1,000 &
Machine translation - Image Captioning \\
\hline
Multi30k &2016 &Flickr30k &German/English &21,014 &186,084 &Machine translation - Image Captioning\\
\hline
Flickr8k-CN &2016 &Flickr8k &Chinese/English &8,000 &45,000 &Image Captioning \\
\hline
AIC-ICC &2017 &Internet &Chinese &240,000 &1,200,000 &Image Captioning \\
\hline
Flickr30k-CN &2017 &Flickr30k &Chinese/English &1,000 &5,000 &Image Captioning \\
\hline
STAIR Captions &2017 &MS-COCO &Japanese/English &164,062 &820,310 &Image Captioning \\
\hline
COCO-CN &2018 &MS-COCO &Chinese/English &20,342 &27,128 &Image tagging - captioning - retrieval \\
\hline
COCO 4K &N/A &MS-COCO &Vietnamese/English &4,000 &20,000 &
Image Captioning \\
\hline
UIT-ViIC &2019 &MS-COCO&Vietnamese/English & 3,850 & 19,250 & Image Captioning \\
    \hline
  \end{tabular}
  
   \label{table:1}
   }
\end{adjustbox}
\end {table*}
We summarize in Table \ref{table:1} an incomplete list of published Image Captioning datasets, in English and in other languages. Several image caption datasets for English have been constructed, the representative examples are Flickr3k \cite{hodosh2013framing}, \cite{rashtchian2010collecting}; Flickr 30k \cite{young2014image} – an extending of Flickr3k and Microsoft COCO (Microsoft Common in Objects in Context) \cite{lin2014microsoft}. 

Besides, several image datasets with non-English captions have been developed. Depending on their applications, the target languages of these datasets vary, including German and French for image retrieval, Japanese for cross-lingual document retrieval \cite{funaki2015image} and image captioning \cite{miyazaki2016cross}, \cite{yoshikawa2017stair}, Chinese for image tagging, captioning and retrieval \cite{li2019coco}. Each of these datasets is built on top of an existing English dataset, with MS-COCO as the most popular choice.

Our dataset UIT-ViIC is constructed using images from Microsoft COCO (MS-COCO). MS-COCO dataset includes more than 150,000 images, divided into three distributions: train, vailidate, test. For each image, five captions are provided independently by Amazon’s Mechanical Turk. MS-COCO is the most popular dataset for Image Captioning thanks to the MS-COCO challenge (2015) and it has a powerful evaluation server for candidates.

Regarding to the Vietnamese language processing, there are quite a number of research works on other tasks such as parsing, part-of-speech, named entity recognition, sentiment analysis, question answering.  However, to the extent of our knowledge, there are no research publications on image captioning for Vietnamese.  Therefore, we decide to build a new corpus of Vietnamese image captioning for Image Captioning research community and evaluate the state-of-the-art models on our corpus. In particular, we validate and compare the results by BLEU \cite{papineni2002bleu}, ROUGE \cite{lin2004rouge} and CIDEr \cite{vedantam2015cider} metrics between Neural Image Captioning (NIC) model \cite{nic}, Image Captioning model from the Pytorch-tutorial \cite{pytorchtutorial} by Yunjey on our corpus as the pioneering results.

\section{Dataset Creation}
\label{dataset}

This section demonstrates how we constructed our new Vietnamese dataset. The dataset consists of 3,850 images relating to sports played with balls from 2017 edition of Microsoft COCO. Similar to most Image Captioning datasets, we provide five Vietnamese captions for each image, summing up to 19,250 captions in total.

\subsection{Annotation Tool with Content Suggestions}
\begin{figure}
    \centering
    \includegraphics[width=\linewidth]{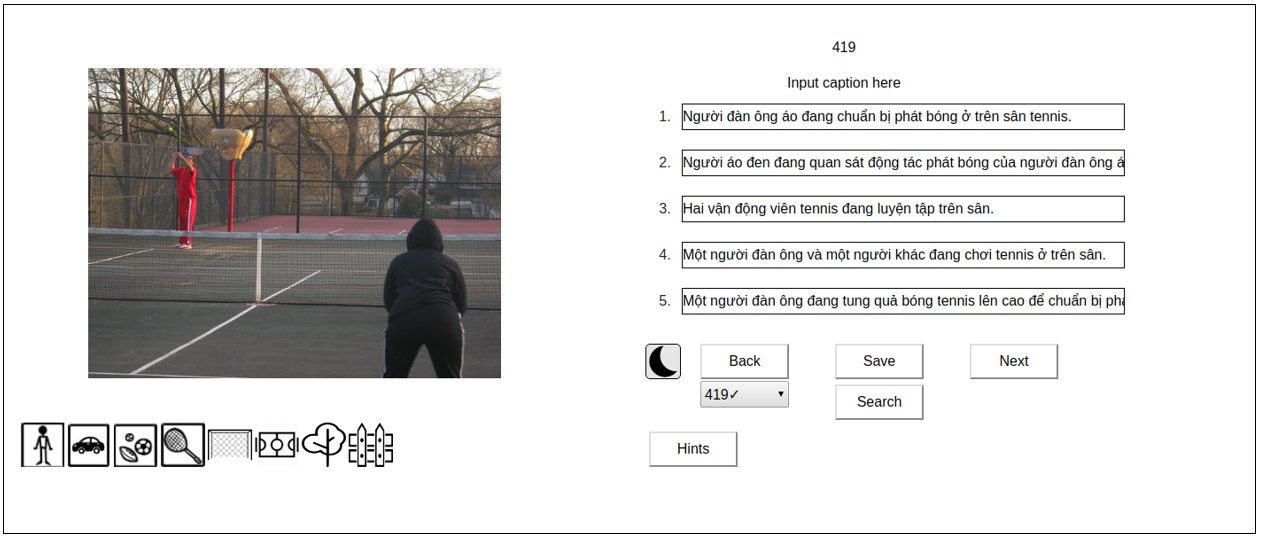}
    \caption{\textbf{Examples of User Interfaces for Image Captions Annotations.}}
    \label{fig:AnnotationTool}
\end{figure}

To enhance annotation efficiency, we present a web-based application for caption annotation. Fig. \ref{fig:AnnotationTool} is the annotation screen of the application.

Our tool assists annotators conveniently load images into a display and store captions they created into a new dataset. With saving function, annotator can save and load written captions for reviewing purposes. Furthermore, users are able to look back their works or the others’ by searching image by image ids.

The tool also supports content suggestions taking advantage of existing information from MS-COCO. First, there are categories hints for each image, displaying as friendly icon. Second, original English captions are displayed if annotator feels their needs. Those content suggestions are helpful for annotators who can’t clearly understand images, especially when there are issues with images’ quality.

\subsection{Annotation Process}

In this section, we describes procedures of building our sportball Vietnamese dataset, called UIT-ViIC.

Our human resources for dataset construction involve five writers, whose ages are from 22-25. Being native Vietnamese residents, they are fluent in Vietnamese. All five UIT-ViIC creators first research and are trained about sports knowledge as well as the specialized vocabulary before starting to work.

During annotation process, there are inconsistencies and disagreements between human's understandings and the way they see images. According to Micah Hodosh et al \cite{hodosh2013framing}, most images’ captions on Internet nowadays tend to introduce information that cannot be obtained from the image itself, such as people name, location name, time, etc. Therefore, to successfully compose meaningful descriptive captions we expect, their should be strict guidelines. 

Inspired from MS-COCO annotation rules \cite{chen2015microsoft}, we first sketched UIT-ViIC's guidelines for our captions:

\begin{enumerate}
	\item Each caption must contain at least ten Vietnamese words.
    \item Only describe visible activities and objects included in image.
    \item Exclude name of places, streets (Chinatown, New York, etc.) and number (apartment numbers, specific time on TV, etc.)
    \item Familiar English words such as laptop, TV, tennis, etc. are allowed.
    \item Each caption must be a single sentence with continuous tense.
    \item Personal opinion and emotion must be excluded while annotating.
    \item Annotators can describe the activities and objects from different perspectives.
    \item Visible “thing” objects are the only one to be described.
    \item Ambiguous “stuff” objects which do not have obvious “border” are ignored.
    \item In case of 10 to 15 objects which are in the same category or species, annotators do not need to include them in the caption.
\end{enumerate}

In comparison with MS-COCO \cite{chen2015microsoft} data collection guidelines in terms of annotation, UIT-ViIC’s guidelines has similar rules (1, 2, 8, 9, 10) . We extend from MS-COCO’s guidelines with five new rules to our own and have modifications in the original ones.

In both datasets, we would like to control sentence length and focus on describing important subjects only in order to make sure that essential information is mainly included in captions. The MS-COCO threshold for sentence’s length is 8, and we raise the number to 10 for our dataset. One reason for this change is that an object in image is usually expressed in many Vietnamese words. For example, a “baseball player” in English can be translated into “vận động viên bóng chày” or “cầu thủ bóng chày”, which already accounted for a significant length of the Vietnamese sentence. In addition, captions must be single sentences with continuous tense as we expect our model’s output to capture what we are seeing in the image in a consise way.

On the other hand, proper name for places, streets, etc must not be mentioned in this dataset in order to avoid confusions and incorrect identification names with the same scenery for output. Besides, annotators’ personal opinion must be excluded for more meaningful captions. Vietnamese words for several English ones such as tennis, pizza, TV, etc are not existed, so annotators could use such familiar words in describing captions. For some images, the subjects are ambiguous and not descriptive which would be difficult for annotators to describe in words. That’s the reason why annotators can describe images from more than one perspective. 

\subsection{Dataset Analysis}

\begin{figure}
    \centering
    \includegraphics[width=\linewidth]{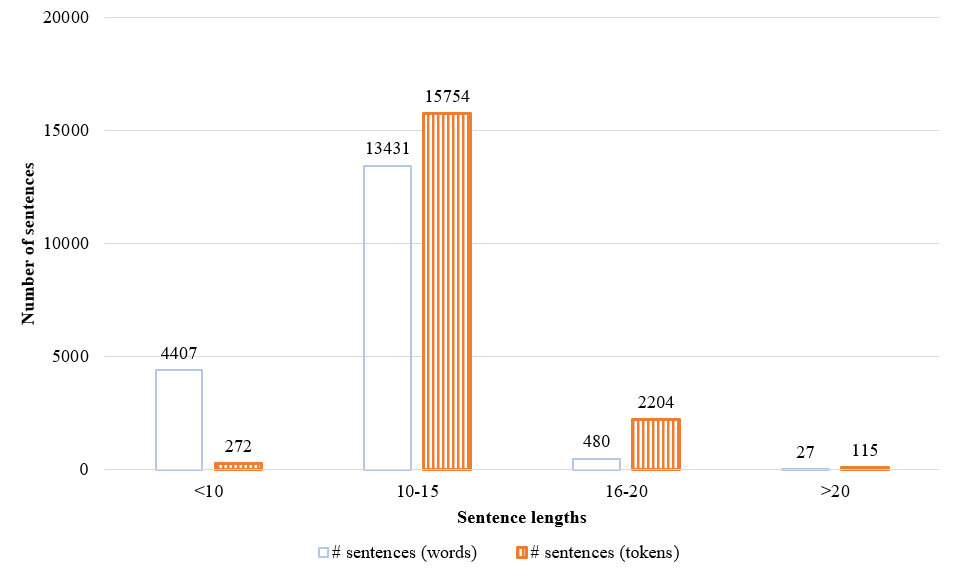}
    \caption{\textbf{Dataset Analysis by Sentences Length.}}
    \label{fig:GeneralDiagram}
\end{figure}

After finishing constructing UIT-ViIC dataset, we have a look in statistical analysis on our corpus in this section. UIT-ViIC covers 3,850 images described by 19,250 Vietnamese captions. Sticking strictly to our annotation guidelines, the majority of our captions are at the length of 10-15 tokens. We are using the term “tokens” here as a Vietnamese word can consist of one, two or even three tokens. Therefore, to apply Vietnamese properly to Image Captioning, we present a tokenization tool - PyVI \cite{pyvi}, which is specialized for Vietnamese language tokenization, at words level. The sentence length using token-level tokenizer and word-level tokenizer are compared and illustrated in Fig.~ \ref{fig:GeneralDiagram}, we can see there are variances there. So that, we can suggest that the tokenizer performs well enough, and we can expect our Image Captioning models to perform better with Vietnamese sentences that are tokenized, as most models perform more efficiently with captions having fewer words.

\begin{table}[H]
\caption{Statistics on types of Vietnamese words.}
   \label{table:2}
\centering

  \begin{tabular}{|c|c|c|}
  \hline
  \textbf{Verbs}&{\textbf{Nouns}}&{\textbf{Adjectives}}\\
\hline
cầm (hold): 3,344 &bóng (ball): 7,686 & tennis: 3,005  \\
\hline
chơi (play): 2,760 &sân (pitch): 6,725 &bóng chày (baseball): 880\\
\hline
đánh bóng (hit): 2,581 &cầu thủ (player): 2,635 &cao (tall): 687\\
\hline
  \end{tabular}
  
\end {table}
Table \ref{table:2} summarizes top three most occuring words for each part-of-speech. Our dataset vocabulary size is 1,472 word classes, including 723 nouns, 567 verbs,  and 182 adjectives. It is no surprise that as our dataset is about sports with balls, the noun “bóng”  (meaning “ball") occurs most, followed by “sân” and "cầu thủ" (“pitch” and “athlete” respectively). We also found that the frequency of word “tennis” stands out among other adjectives, which specifies that the set covers the majority of tennis sport, followed by “bóng chày” (meaning “baseball”). Therefore, we expect our model to generate the best results for tennis images.

\section{Image Captioning Models}
\label{models}
Our main goal in this section is to see if Image Captioning models could learn well with  Vietnamese language.  To accomplish this task, we train and evaluate our dataset with two published Image Captioning models applying encoder-decoder architecture. The models we propose are Neural Image Captioning (NIC) model \cite{nic},  Image Captioning model from the Pytorch-tutorial \cite{pytorchtutorial} by Yunjey.

Overall, CNN is first used for extracting image features for encoder part. The image features which are presented in vectors will be used as layers for decoding. For decoder part, RNN - LSTM are used to embed the vectors to target sentences using words/tokens provided in vocabulary.

\subsection{Model from Pytorch tutorial}

Model from pytorch-tutorial by Yunjey applies the baseline technique of CNN and LSTM for encoding and decoding images. Resnet-152 \cite{he2016deep} architecture is proposed for encoder part, and we use the pretrained one on ILSVRC-2012-CLS \cite{russakovsky2015imagenet} image classification dataset to tackle our current problem. LSTM is then used in this model to generate sentence word by word.

\subsection{NIC - Show and tell model}

NIC - Show and Tell uses CNN model which is currently yielding the state-of-the-art results. The model achieved 0.628 when evaluating on BLEU-1 on COCO-2014 dataset. For CNN part, we utilize VGG-16 \cite{simonyan2014very} architecture pre-trained on COCO-2014 image sets with all categories. In decoding part, LSTM is not only trained to predict sentence but also to compute probability for each word to be generated. As a result, output sentence will be chosen using search algorithms to find the one that have words yielding the maximum probabilities.

\section{Experiments}
\label{experiment}
\subsection{Experiment Settings}
\subsubsection{Dataset preprocessing}

As the images in our dataset are manually annotated by human, there are mistakes including grammar, spelling or extra spaces, punctuation. Sometimes, the Vietnamese’s accent signs are placed in the wrong place due to distinct keyboard input methods. Therefore, we eliminate those common errors before working on evaluating our models.

\subsubsection{Dataset preparation}

We conduct our experiments and do comparisons through three datasets with the same size and images of sportball category: Two Vietnamese datasets generated by two methods (translated by Google Translation service and annotated by human) and the original MS-COCO English dataset. The three sets are distributed into three subsets: 2,695 images for the training set, 924 images for validation set and 231 images for test set.

\subsection{Evaluation Measures}

To evaluate our dataset, we use metrics proposed by most authors in related works of extending Image Captioning dataset, which are BLEU \cite{papineni2002bleu}, ROUGE \cite{lin2004rouge} and CIDEr \cite{vedantam2015cider}. BLEU and ROUGE are often used mainly for text summarization and machine translation, whereas CIDEr was designed especially for evaluating Image Captioning models.

\subsubsection{Comparison methods}

We do comparisons with three sportball datasets, as follows:
\begin{itemize}
\item \textbf{Original English (English-sportball)}: The original MS-COCO English dataset with 3,850 sportball images. This dataset is first evaluated in order to have base results for following comparisons.
\item \textbf{Google-translated Vietnamese (GT-sportball)}: The translated MS-COCO English dataset into Vietnamese using Google Translation API, categorized into sportball.
\item \textbf{Manually-annotated Vietnamese (UIT-ViIC)}: The Vietnamese dataset built with manually written captions for images from MS-COCO, categorized into sportball.
\end{itemize}
\subsection{Experiment Results}

The two following tables, Table~\ref{table:3} and Table~\ref{table:4},  summarize experimental results of Pytorch-tutorial, NIC - Show and Tell models. The two models are trained with three mentioned datasets, which are English-sportball, GT-sportball, UIT-ViIC. After training, 924 images from validation subset for each dataset are used to validate the our models.

\begin{table}
\caption{Experimental results of pytorch-tutorial models}
\label{table:3}
\centering
\begin{tabular}{ |c|c|r|r|r|r|r|r|}
\hline
Dataset& Tokenizer&BLEU-1&BLEU-2&BLEU-3&BLEU-4&ROUGE-L&CIDEr-D \\
\hline
English-sportball&nltk&\textbf{0.761}&0.562&0.405&0.289&0.560&0.668 \\
\hline
GT-sportball&PyVI&0.596&0.455&0.341&0.254&0.522&0.578 \\
\hline
UIT-ViIC&PyVI&0.710&0.575&0.476&0.394&0.626&\textbf{1.005} \\
\hline
\end{tabular}

\vspace{2em}

\caption{Experimental results of NIC - Show and Tell models}
\label{table:4}
\centering
\begin{tabular}{ |c|c|r|r|r|r|r|r|}
\hline
Dataset& Tokenizer&BLEU-1&BLEU-2&BLEU-3&BLEU-4&ROUGE-L&CIDEr-D \\
\hline
English-sportball&nltk&\textbf{0.689}&0.501&0.355&0.252&0.585&0.667 \\
\hline
GT-sportball&PyVI&0.643&0.481&0.368&0.281&0.565&0.567 \\
\hline
UIT-ViIC&PyVI&0.682&\textbf{0.561}&\textbf{0.411}&\textbf{0.327}&0.599&\textbf{0.818} \\
\hline
\end{tabular}
\vspace{-1em}
\end{table}

As can be seen in Table~\ref{table:3}, with model from Pytorch tutorial, MS-COCO English captions categorized with sportball yields better results than the two Vietnamese datasets. However, as number of consecutive words considered (BLEU gram) increase, UIT-ViIC’s BLEU scores start to pass that of English sportball and their gaps keep growing. The ROUGE-L and CIDEr-D scores for UIT-ViIC model prove the same thing, and interestingly, we can observe that the CIDEr-D score for the UIT-ViIC model surpasses English-sportball counterpart.

The same conclusion can be said from Table~\ref{table:4}. Show and Tell model’s results show that MS-COCO sportball English captions only gives better result at BLEU-1. From BLEU-3 to BLEU-4, both GT-sportball and UIT-ViIC yield superior scores to English-sportball. Besides, when limiting MS-COCO English dataset to sportball category only, the results are higher (0.689, 0.501, 0.355, 0.252) than when the model is trained on MS-COCO with all images, which scored only 0.629, 0.436, 0.290, 0.193 (results without tuning in 2018) from BLEU-1 to BLEU-4 respectively.

When we compare between two Vietnamese datasets, UIT-ViIC models perform better than sportball dataset translated automatically, GT-sportball. The gaps between the two results sets are more trivial in NIC model, and the numbers get smaller as the BLEU’s n-gram increase.

\begin{figure}
    \centering
    \includegraphics[width=\linewidth]{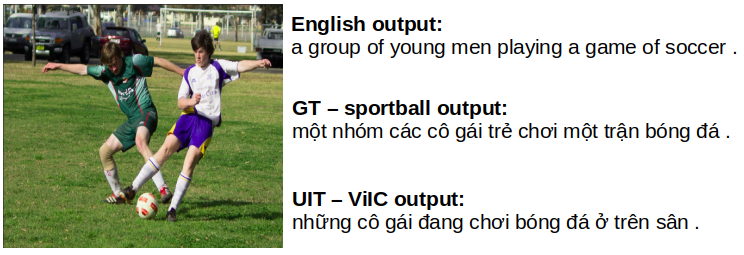}
    \includegraphics[width=\linewidth]{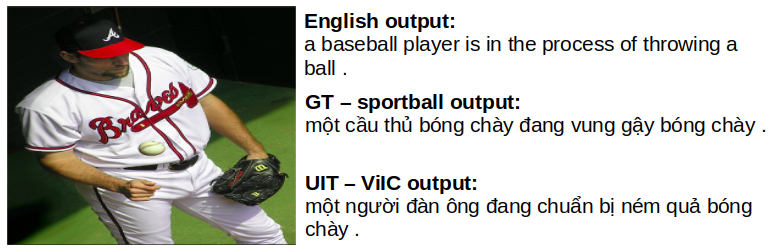}
    \caption{\textbf{Examples of captions generated by models from pytorch-tutorial trained on the three datasets that yieled expected outputs.}}
    \label{fig:ExpectedCaption}
    \vspace{-1em}
\end{figure}

In Fig.~\ref{fig:ExpectedCaption}, two images inputted into the models generate two Vietnamese captions that are able to describe accurately the sport game, which is soccer. The two models can also differentiate if there is more than one person in the images. However, when comparing GT-sportball outputs with UIT-ViIC ones in both images, UIT-ViIC yield captions that sound more naturally, considering Vietnamese language. Furthermore, UIT-ViIC demonstrates the specific action of the sport more accurately than GT-sportball. For example, in the below image of Fig.~\ref{fig:ExpectedCaption}, UIT-ViIC tells the exact action (the man is preparing to throw the ball), whereas GT-sportball is mistaken (the man swing the bat). The confusion of GT-sportball happens due to GT-sportball train set is translated from original MS-COCO dataset, which is annotated in more various perspective and wider vocabulary range with the dataset size is not big enough.


There are cases when the main objects are too small, both English and GT - sportball captions tell the unexpected sport, which is tennis instead of baseball, for instance. Nevertheless, the majority of UIT-ViIC captions can tell the correct type of sport and action, even though the gender and age identifications still need to be improved.

\section{Conclusion and Further Improvements}
\label{conclusion}

In this paper, we constructed a Vietnamese dataset with images from MS-COCO, relating to the category within sportball, consisting of 3,850 images with 19,250 manually-written Vietnamese captions. Next, we conducted several experiments on two popular existed Image Captioning models to evaluate their efficiency when learning two Vietnamese datasets. The results are then compared with the original MS-COCO English categorized with sportball category. 

Overall, we can see that English set only out-performed Vietnamese ones in BLEU-1 metric, rather, the Vietnamese sets performing well basing on BLEU-2 to BLEU-4, especially CIDEr scores. On the other hand, when UIT-ViIC is compared with the dataset having captions translated by Google, the evaluation results and the output examples suggest that Google Translation service is able to perform acceptablly even though most translated captions are not perfectly natural and linguistically friendly. As a results, we proved that manually written captions for Vietnamese dataset is currently prefered.

For future improvements, extending the UIT-ViIC's cateogry into all types of sport to verify how the dataset's size and category affect the Image Captioning models' performance is considered as our highest priority. Moreover, the human resources for dataset construction will be expanded. Second, we will continue to finetune our experiments to find out proper parameters for models, especially with encoding and decoding architectures, for better learning performance with Vietnamese dataset, especially when the categories are limited.\\

%
%

%
%
%

\begin{thebibliography}{8}
\bibitem{chen2015microsoft}
Chen, X., Fang, H., Lin, T.Y., Vedantam, R., Gupta, S., Dollár, P. and Zitnick, C.L., 2015. Microsoft coco captions: Data collection and evaluation server. arXiv preprint arXiv:1504.00325.


\bibitem{Donahue_2015_CVPR}
Donahue, J., Anne Hendricks, L., Guadarrama, S., Rohrbach, M., Venugopalan, S., Saenko, K. and Darrell, T., 2015. Long-term recurrent convolutional networks for visual recognition and description. In Proceedings of the IEEE conference on computer vision and pattern recognition (pp. 2625-2634).

\bibitem{horus2017}
Eyra, Horus, https://italianinnovationday.weebly.com/horus-technology.html

\bibitem{funaki2015image}
Funaki, R. and Nakayama, H., 2015, September. Image-mediated learning for zero-shot cross-lingual document retrieval. In Proceedings of the 2015 Conference on Empirical Methods in Natural Language Processing (pp. 585-590).

\bibitem{gao2017video}
Gao, L., Guo, Z., Zhang, H., Xu, X. and Shen, H.T., 2017. Video captioning with attention-based LSTM and semantic consistency. IEEE Transactions on Multimedia, 19(9), pp.2045-2055.

\bibitem{he2016deep}
He, K., Zhang, X., Ren, S. and Sun, J., 2016. Deep residual learning for image recognition. In Proceedings of the IEEE conference on computer vision and pattern recognition (pp. 770-778).

\bibitem{hodosh2013framing}
Hodosh, M., Young, P. and Hockenmaier, J., 2013. Framing image description as a ranking task: Data, models and evaluation metrics. Journal of Artificial Intelligence Research, 47, pp.853-899.

\bibitem{hossain2019comprehensive}
Hossain, M.D., Sohel, F., Shiratuddin, M.F. and Laga, H., 2019. A comprehensive survey of deep learning for image captioning. ACM Computing Surveys (CSUR), 51(6), p.118.

\bibitem{karpathy2015deep}
Karpathy, A. and Fei-Fei, L., 2015. Deep visual-semantic alignments for generating image descriptions. In Proceedings of the IEEE conference on computer vision and pattern recognition (pp. 3128-3137).

\bibitem{li2019coco}
Li, X., Xu, C., Wang, X., Lan, W., Jia, Z., Yang, G. and Xu, J., 2019. COCO-CN for Cross-Lingual Image Tagging, Captioning and Retrieval. IEEE Transactions on Multimedia.

\bibitem{lin2004rouge}
Lin, C.Y., 2004. Rouge: A package for automatic evaluation of summaries. In Text summarization branches out (pp. 74-81).

\bibitem{lin2014microsoft}
Lin, T.Y., Maire, M., Belongie, S., Hays, J., Perona, P., Ramanan, D., Dollár, P. and Zitnick, C.L., 2014, September. Microsoft coco: Common objects in context. In European conference on computer vision (pp. 740-755). Springer, Cham.

\bibitem{miyazaki2016cross}
Miyazaki, T. and Shimizu, N., 2016, August. Cross-lingual image caption generation. In Proceedings of the 54th Annual Meeting of the Association for Computational Linguistics (Volume 1: Long Papers) (pp. 1780-1790).

\bibitem{nic}
Octavio Arriaga, Neural image captioning (NIC), https://github.com/oarriaga/neural\_image\_captioning

\bibitem{papineni2002bleu}
Papineni, K., Roukos, S., Ward, T. and Zhu, W.J., 2002, July. BLEU: a method for automatic evaluation of machine translation. In Proceedings of the 40th annual meeting on association for computational linguistics (pp. 311-318). Association for Computational Linguistics.

\bibitem{rashtchian2010collecting}
Rashtchian, C., Young, P., Hodosh, M. and Hockenmaier, J., 2010, June. Collecting image annotations using Amazon's Mechanical Turk. In Proceedings of the NAACL HLT 2010 Workshop on Creating Speech and Language Data with Amazon's Mechanical Turk (pp. 139-147). Association for Computational Linguistics.

\bibitem{russakovsky2015imagenet}
Russakovsky, O., Deng, J., Su, H., Krause, J., Satheesh, S., Ma, S., Huang, Z., Karpathy, A., Khosla, A., Bernstein, M. and Berg, A.C., 2015. Imagenet large scale visual recognition challenge. International journal of computer vision, 115(3), pp.211-252.

\bibitem{simonyan2014very}
Simonyan, K. and Zisserman, A., 2014. Very deep convolutional networks for large-scale image recognition. arXiv preprint arXiv:1409.1556.

\bibitem{staniute2019systematic}
Staniūtė, R. and Šešok, D., 2019. A Systematic Literature Review on Image Captioning. Applied Sciences, 9(10), p.2024.

\bibitem{vedantam2015cider}
Vedantam, R., Lawrence Zitnick, C. and Parikh, D., 2015. Cider: Consensus-based image description evaluation. In Proceedings of the IEEE conference on computer vision and pattern recognition (pp. 4566-4575).

\bibitem{pyvi}
Viet Trung Tran, Python Vietnamese Core NLP Toolkit, https://github.com/trungtv/pyvi

\bibitem{vinyals2015show}
Vinyals, O., Toshev, A., Bengio, S. and Erhan, D., 2015. Show and tell: A neural image caption generator. In Proceedings of the IEEE conference on computer vision and pattern recognition (pp. 3156-3164).

\bibitem{pytorchtutorial}
Y. Choi, Pytorch Tutorial: Deep Convolutional GAN, https://github.com/pytorch/pytorch

\bibitem{yoshikawa2017stair}
Yoshikawa, Y., Shigeto, Y. and Takeuchi, A., 2017. Stair captions: Constructing a large-scale japanese image caption dataset. arXiv preprint arXiv:1705.00823.

\bibitem{young2014image}
Young, P., Lai, A., Hodosh, M. and Hockenmaier, J., 2014. From image descriptions to visual denotations: New similarity metrics for semantic inference over event descriptions. Transactions of the Association for Computational Linguistics, 2, pp.67-78.



\end{thebibliography}
%

\end{document}